\documentclass[final,5p,times,twocolumn,a4paper,numbers,nopreprintline]{elsarticle}

\usepackage[utf8]{inputenc}
\usepackage[T1]{fontenc}
\usepackage{microtype}

\usepackage{amsmath}
\usepackage{amssymb}
\usepackage{amsfonts}
\usepackage{bm}
\usepackage{siunitx}
\graphicspath{{img/}}
\usepackage{booktabs}
\usepackage{multirow}
\usepackage{subcaption}
\usepackage{array}
\usepackage{tabularx}
\usepackage{float}
\usepackage{placeins}
\usepackage[font=small,labelfont=bf]{caption}
\usepackage{algorithm}
\usepackage{algpseudocode}
\usepackage{enumitem}
\usepackage{chngcntr}

\biboptions{sort&compress}
\usepackage[hidelinks]{hyperref}
\usepackage{cleveref}


\newcommand{\real}{\mathbb{R}}
\newcommand{\transpose}{^{\top}}
\newcommand{\tidx}{t}
\newcommand{\horizon}{T}

\newcommand{\observation}{\bm{O}}
\newcommand{\discount}{\gamma}

\newcommand{\cartyaw}{\theta^{\mathrm{cart}}}

\newcommand{\cartcmd}{\bm{v}^{\mathrm{cart,cmd}}}
\newcommand{\carttwist}{\bm{v}^{\mathrm{cart}}}
\newcommand{\cmdvel}{\bm{v}^{\mathrm{cmd}}}
\newcommand{\policygmt}{\pi_{\mathrm{base}}}
\newcommand{\policytask}{\pi_{\mathrm{task}}}

\newcommand{\resaction}{\bm{a}^{\mathrm{arm}}}

\newcommand{\wrapangle}{\operatorname{wrap}}
\newcommand{\errlat}{\varepsilon}

\newcommand{\envwrench}{\bm{F}^{\mathrm{ext}}}

\newcommand{\legaction}{\bm{a}^{\mathrm{loco}}}

\newcommand{\basevel}{\bm{v}^{\mathrm b}}

\newcommand{\handwrenchest}{\hat{\bm{F}}}

\newcommand{\motionrmse}{\mathrm{RMSE}}
\newcommand{\metricvelocity}{v}
\newcommand{\metriccommand}{v^{\mathrm{cmd}}}
\newcommand{\metricerror}{e}
\newcommand{\metricmean}[1]{\mu_{#1}}
\newcommand{\motionripple}{\mathrm{RMSF}}

\newcommand{\cmdres}{\bm{a}^{\mathrm{base}}}
\newcommand{\rewloco}{r^{\mathrm{loco}}}
\newcommand{\locotrack}{r^{\mathrm{track}}}
\newcommand{\locobalance}{r^{\mathrm{balance}}}
\newcommand{\locogait}{r^{\mathrm{gait}}}
\newcommand{\locoregcost}{c^{\mathrm{reg}}}
\newcommand{\locotrackwidth}{\sigma}
\newcommand{\taskreward}{r^{\mathrm{task}}}
\newcommand{\tasksmoothcost}{c^{\mathrm{smooth}}}
\newcommand{\tasksupport}{r^{\mathrm{support}}}

\newcommand{\handwrenchlabel}{\bm{F}^{\star}}

\newcommand{\worldyaw}{\psi^{\mathrm{cart}}}
\newcommand{\pathheading}{\psi^{\mathrm{ref}}}

\newcommand{\rewweight}[1]{w_{#1}}

\newcommand{\drivechannel}{\bm{d}^{\mathrm{dr}}}
\newcommand{\drivemap}{\bm{U}}

\newcommand{\capweight}{\bm{\Sigma}}

\newcommand{\intwrench}{\bm{F}^{\mathrm{int}}}

\newcommand{\costint}{J^{\mathrm{int}}}
\newcommand{\planarsel}{\bm{P}}
\newcommand{\intratio}{\rho^{\mathrm{int}}}
\newcommand{\estimatedsignal}{\hat{y}}
\newcommand{\referencesignal}{y}
\newcommand{\referencemean}{\bar{y}}

\newcommand{\fullaction}{\bm{a}^{\mathrm{res}}}
\newcommand{\legresout}{\bm{a}^{\mathrm{leg}}}

\newcommand{\sysmodel}{\mathcal{M}}
\newcommand{\locomodel}{\mathcal{M}_1}
\newcommand{\pushmodel}{\mathcal{M}_2}
\newcommand{\sysstate}{\bm{s}}
\newcommand{\obsstate}{\bm{s}^{\mathrm{obs}}}
\newcommand{\unobsstate}{\bm{s}^{\mathrm{unobs}}}
\newcommand{\robotobs}{\bm{s}^{\mathrm{obs,r}}}
\newcommand{\wholeaction}{\bm{a}}
\newcommand{\handwrench}{\bm{F}^{\mathrm{hand}}}

\newcommand{\unobsest}{\mathcal{E}}
\newcommand{\unobshat}{\hat{\bm{s}}^{\mathrm{unobs}}}

\newcommand{\teacherforce}{\bm{z}^{F}}
\newcommand{\teacherstate}{\bm{z}^{s}}
\newcommand{\privstate}{\bm{e}}
\newcommand{\teacherforceenc}{E_F^{T}}
\newcommand{\teacherstateenc}{E_s^{T}}
\newcommand{\interactionhistory}{\bm{H}}
\newcommand{\forcelatentseq}{\hat{\bm{Z}}^{F}}
\newcommand{\studentforce}{\hat{\bm{z}}^{F}}
\newcommand{\studentstate}{\hat{\bm{z}}^{s}}
\newcommand{\forcedecoder}{D_F}
\newcommand{\studentforceenc}{E_F^{S}}
\newcommand{\interactionfeatures}{\bm{h}}
\newcommand{\interactionfusion}{G}
\newcommand{\statehead}{D_s}
\newcommand{\motionvelest}{\hat{\bm{v}}}
\newcommand{\motionaccest}{\hat{\bm{a}}}
\newcommand{\lossstudent}{\mathcal{L}_{\mathrm{student}}}
\newcommand{\lossalign}{\mathcal{L}_{\mathrm{align}}}
\newcommand{\lossforce}{\mathcal{L}_{F}}
\newcommand{\lossmotionvel}{\mathcal{L}_{v}}
\newcommand{\lossmotionacc}{\mathcal{L}_{a}}
\newcommand{\stopgrad}{\operatorname{sg}}
\newcommand{\estlossweight}[1]{\lambda_{#1}}
\newcommand{\carttrack}{r^{\mathrm{cart}}}
\newcommand{\graspcost}{c^{\mathrm{grasp}}}

\newproof{pf}{Proof}

\usepackage{xcolor}

\journal{Chinese Journal of Mechanical Engineering}

\newcommand{\figureplaceholder}[3]{%
  \IfFileExists{img/#1}{\includegraphics[width=\linewidth]{img/#1}}{%
    \fbox{\begin{minipage}[c][#2][c]{0.94\linewidth}
      \centering\small\color{blue!65!black}
      #3\par\medskip
    \end{minipage}}}}
\begin{document}
\begin{frontmatter}
\title{ForPC: Force-Aware Whole-Body Control for Precise Humanoid Cart Pushing}

\author{ZhenYu Zou}
\author{Xing Liu}
\author{ShiChuang Wang}
\author{YunLong Dong}
\author{LiNan Deng}
\author{Feng Hua}
\author{Jun Wan}
\author{Cheng Cheng\textsuperscript{*}}

\patchcmd{\MaketitleBox}
  {\elsauthors\par\vskip10pt}
  {\elsauthors\par\vskip10pt
   {\normalsize\normalfont Project Website: \url{https://roboforpc.github.io/}}\par}
  {}{\PackageError{CJME}{Unable to place project website below authors}{Check the elsarticle title block.}}

\begin{abstract}
Humanoid robots are expected to take over strenuous material handling in manufacturing, logistics, and healthcare, where operating existing wheeled carts would let them move heavy loads without modifying the workplace.
However, humanoid cart pushing remains challenging because the bimanual grasp closes a force-coupled loop in which the contact wrenches that drive the cart also disturb the robot's balance, and payload changes alter the coupled dynamics.
We propose ForPC, a force-aware whole-body control framework that regulates cart velocity through two-stage reinforcement learning.
A locomotion prior is first trained under varied hand loads and upper-body postures to provide balance support.
A force latent and a state latent are then inferred from onboard observations, without external force sensors, and fed to a residual policy that coordinates bounded arm, waist, and leg adjustments with base-velocity commands.
Multi-objective rewards penalize bimanual internal opposition, grasp slip, and actuator saturation, encouraging the policy to apply contact force on demand without human demonstrations.
Sim-to-sim experiments show that ForPC reduces cart-velocity RMSE by up to 68.2\% relative to a locomotion-only baseline (Loco-only) and cross-track RMSE by up to 74.2\% on the multi-bend route.
Trained solely in simulation, the policy also transfers to a physical Unitree G1 without real-world training or fine-tuning and, with no payload-specific adjustment, pushes a cart carrying up to \SI{30}{kg} at \SI{0.5}{m/s}.
These results suggest that humanoid robots could eventually replace humans for efficient material handling in daily life.
\end{abstract}
\apptocmd{\MaketitleBox}{%
  \par\noindent\begin{minipage}{\textwidth}
    \centering
    \captionsetup{type=figure}
    \includegraphics[width=\linewidth]{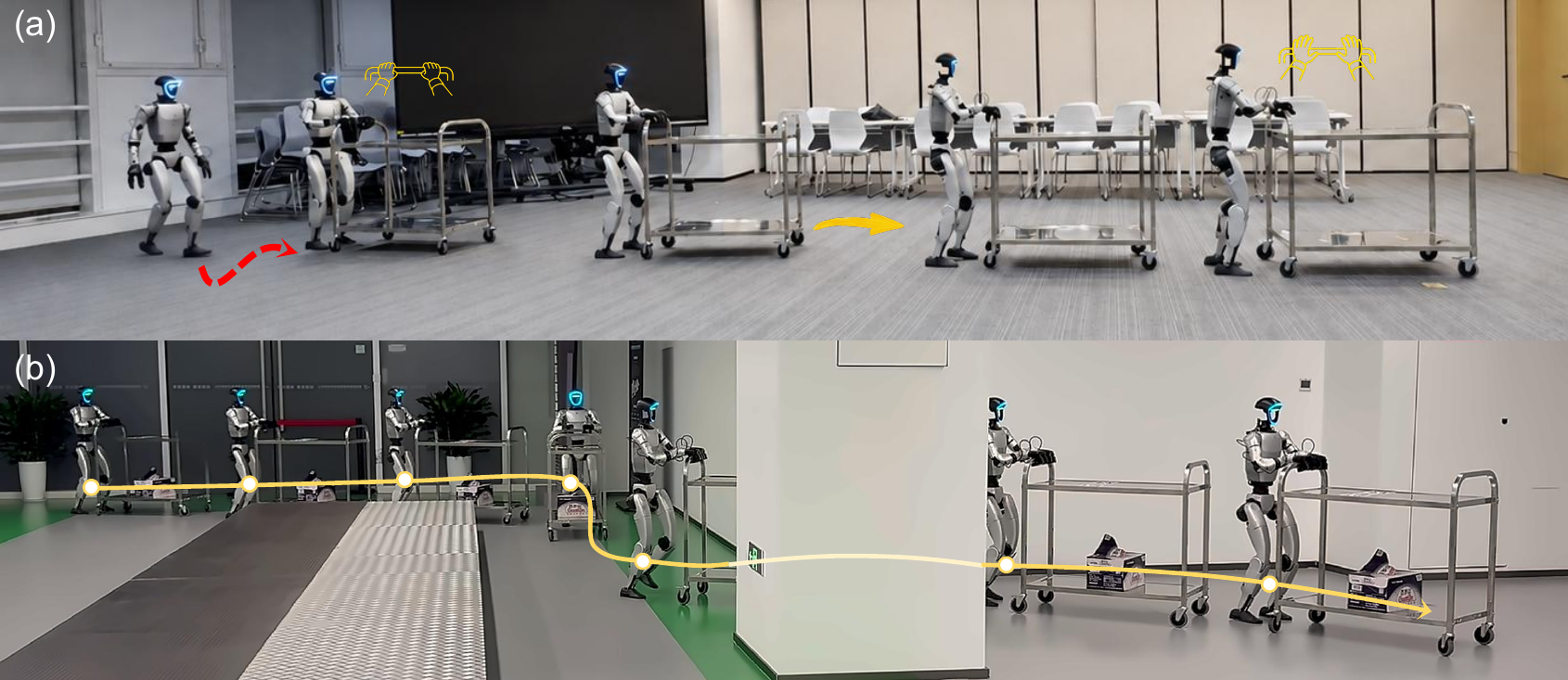}
    \caption{(a) Complete cart-pushing sequence performed by a Unitree G1 humanoid, comprising cart grasping, pushing, and release.
    (b) A Unitree G1 humanoid pushing the cart along a complex path.
    }
    \label{fig:cart_teaser}
  \end{minipage}\par\vspace{10pt}
}{}{\PackageError{CJME}{Unable to place Figure 1 below the abstract}{Check the elsarticle title block.}}
\end{frontmatter}

\section{Introduction}
\label{sec:introduction}

Humanoid robots possess an anthropomorphic morphology and whole-body dexterity, enabling them to directly operate human-centric equipment, including manual wheeled carts designed for human workers~\citep{tong2024humanoid,gu2026humanoid,kheddar2019humanoid}.
By autonomously taking over repetitive and strenuous hauling tasks, humanoid robots could enhance operational productivity and transport efficiency, and thus hold broad application prospects.

Despite early progress in humanoid cart transport~\citep{hawley2017rolling,rioux2018autonomous}, achieving precise and dependable pushing remains challenging.
Sustained contact establishes a bilaterally coupled physical system, in which hand forces drive the cart while its motion inertia and rolling resistance transmit reaction loads back to the robot and disrupt its locomotion balance.
Under payload variations, the robot must accurately regulate both the magnitude and the direction of its contact effort to maintain a steady heading and forward cruise.
Grasping the same handle makes uncoordinated arms prone to working against each other.
This mutual opposition produces internal forces that do not contribute to propulsion~\citep{yoshikawa1991manipulating} and consumes the joint torque margin needed for heading corrections and steering maneuvers.

Model-based formulations coordinate motion and force tasks with balance objectives but require explicit models of the contact and the object~\citep{sentis2005synthesis,wensing2024optimization,sleiman2021unified}, whereas learning-based methods obtain whole-body coordination from simulated interaction, demonstrations, or motion tracking~\citep{dao2024sim2real,ze2025twist,zhao2025resmimic,liao2026beyondmimic}.
Recent controllers further accommodate manipulation loads~\citep{ben2025homie,zhang2026falcon,xiao2025kinematics} or infer contact forces and object motion from the robot's own signals~\citep{portela2024force,zhi2025unified,li2026haic,liu2026dexanytwist,wan2025dexremoe}.
However, Portela et al.~\citep{portela2024force} and Zhi et al.~\citep{zhi2025unified} estimate contact forces without object motion, whereas HAIC~\citep{li2026haic} predicts object motion without an explicit hand-force estimate, so none of them closes the feedback loop on both the cart velocity and the bimanual hand wrenches.
Without such feedback, a policy can neither assess the cart's actual dynamic response nor detect whether the two arms are fighting against each other on the handle.
Although wrist force/torque sensors could provide direct feedback, such sensing is not always available for control, so recent methods estimate these forces from proprioception~\citep{portela2024force,zeng2026forceaware}.
Precise cart pushing under varying payloads, which requires feedback on both the cart motion and the hand forces without external force sensors, therefore remains an unresolved challenge.

Addressing this challenge requires treating the humanoid and the cart as a tightly coupled dynamic system rather than decoupled motion primitives.
Effective payload transport depends on closing the feedback loop around the physical interaction, so the controller must perceive both the cart's real-time motion response and the contact loads at the hands.
Because everyday carts have no built-in motion sensors and the hands provide no direct force measurement, these quantities must be inferred from signals available on the robot itself.

We propose ForPC, a force-aware whole-body control framework that uses two-stage reinforcement learning for dynamic cart pushing, as shown in \Cref{fig:framework}.
In the first stage, an interaction-resilient locomotion prior is pre-trained under severe, randomized external hand disturbances and bounded posture offsets, and is then frozen so that the task policy can reuse its support response.
In the second stage, a teacher--student estimator infers force and state latents from histories of joint kinematics, upper-body torques, and the visually observed relative cart heading, and the task policy uses these latents to output bounded waist--arm and leg joint residuals and modulate robot base velocity commands.
Without relying on external force sensors or human demonstrations, our framework effectively coordinates bimanual effort and transfers zero-shot across varied payloads.
We evaluate ForPC against a locomotion-only baseline (Loco-only) in Isaac Sim and MuJoCo on velocity tracking, force estimation, and route following, and then deploy the policy, trained solely in Isaac Sim, zero-shot on a physical Unitree G1 humanoid to accomplish the transport of parts and workpieces along complex paths (\Cref{fig:cart_teaser}).

The main contributions of this paper are threefold.
\begin{itemize}[leftmargin=*,itemsep=3pt]
\item We propose ForPC, a two-stage whole-body control framework for coupled humanoid cart pushing that combines a locomotion prior, interaction estimators, and a residual policy.
%
\item We design a force-guided interaction inference module that infers a force latent from onboard signals and uses it to guide the estimation of a state latent, allowing the policy to perceive contact loading and cart motion without external force sensors.
%
\item We transfer our policy zero-shot to a real Unitree G1 humanoid, which achieves precise cart-pushing control along complex paths under a \SI{30}{kg} payload.
\end{itemize}

\section{Related Work}
\label{sec:related_work}

\subsection{Humanoid Pushing and Object Transport}
Early humanoid pushing relied on dual-arm force control~\citep{nozawa2012pushing}, controllers designed for a rolling cart~\citep{hawley2017rolling}, and joint planning of centre-of-mass motion and manipulation forces~\citep{murooka2021locomanip}, which require force measurements or explicit object models.
Learning-based systems have since substantially increased the load that a humanoid can move.
Xiao et al.~\citep{xiao2025kinematics} push and pull a cart with a total load of \SI{112.8}{kg} under commanded velocities and forces, Li et al.~\citep{li2026mpcrl} use MPC-guided rewards to push a cart payload of up to \SI{290}{kg} without runtime object feedback, and HULK~\citep{dang2026hulk} performs cart pushing and pulling at \SI{150}{kg}.
HAIC~\citep{li2026haic} predicts object motion from proprioception and reference motions to push and pull a cart, WholeBodyVLA~\citep{jiang2026wholebodyvla} pushes a \SI{50}{kg} cart within composite tasks, and other systems address visual goal pushing~\citep{hu2026vofa} and compliant transport~\citep{zhang2026interaction}.
However, these policies rarely estimate the cart motion and the hand forces together, although precise pushing depends on both.
Payload mass and distribution change between transport tasks, so the mapping from hand forces to cart motion varies and is difficult to capture with a fixed object model.
A controller must therefore adjust the magnitude and direction of its hand forces from online feedback.

\subsection{Force Estimation and Force-Adaptive Control for Humanoid Robots}
Recent humanoid controllers increasingly treat contact force as part of the control objective.
FACET~\citep{xu2025facet}, SoftMimic~\citep{margolis2025softmimic}, and GentleHumanoid~\citep{lu2026gentlehumanoid} obtain compliant responses through a virtual mass-spring-damper system, augmented motion data, and impedance-based motion tracking, respectively.
For intense interaction, FALCON~\citep{zhang2026falcon} trains upper- and lower-body agents with a force curriculum, HAFO~\citep{dong2025hafo} couples them through a constrained residual action space, and Thor~\citep{li2025thor} adds a force-adaptive torso-tilt objective.
Much of the sensorless force control originates from other legged platforms, whose estimation principles also apply to humanoids.
Portela et al.~\citep{portela2024force} track commanded end-effector forces from proprioception on a quadruped manipulator, Zhi et al.~\citep{zhi2025unified} unify position and force control on quadrupedal and humanoid platforms, Zeng et al.~\citep{zeng2026forceaware} supply a hybrid force estimate as an explicit policy observation on a wheeled-legged robot, and CompliantWBC~\citep{do2026compliantwbc} brings force latent estimation to a heavy humanoid.
However, these controllers target a commanded force, a compliant response, or disturbance rejection, and they generally do not estimate how the manipulated object moves.

\subsection{Residual Reinforcement Learning for Humanoid Control}
General motion-tracking and locomotion policies provide reusable whole-body skills~\citep{he2024h2o,he2025omnih2o,fu2025humanplus,ze2025twist,liao2026beyondmimic,ben2025homie}, but they lack the precision and object awareness that loco-manipulation requires.
Residual learning addresses this gap by refining an existing controller instead of learning the task from scratch~\citep{johannink2019residual}.
Residual actions compensate for the dynamics mismatch of retargeted references~\citep{sun2026robotdancing} and adapt general trackers to new interfaces~\citep{sun2026mosaic}.
ResMimic~\citep{zhao2025resmimic} refines a general motion-tracking policy with an object-aware residual policy, and HDMI~\citep{weng2025hdmi} co-tracks robot and object states from human videos through a residual action space.
SEEC~\citep{jang2025seec} and SteadyTray~\citep{huang2026steadytray} instead cancel locomotion-induced disturbances at the end effectors and the tray, respectively, and Xiao et al.~\citep{xiao2025kinematics} add a delta-command policy over separately trained upper- and lower-body policies.
However, the object-aware residuals depend on per-task reference motions and often on motion-capture object states, whereas the stabilizing residuals keep the end effectors or payload steady rather than drive an object.
Neither suits sustained interaction with a moving cart, whose route has no reference motion and whose velocity is not measured onboard.

\section{Problem Formulation}
\label{sec:problem}
\begin{figure}[!t]
\centering
\includegraphics[width=\linewidth]{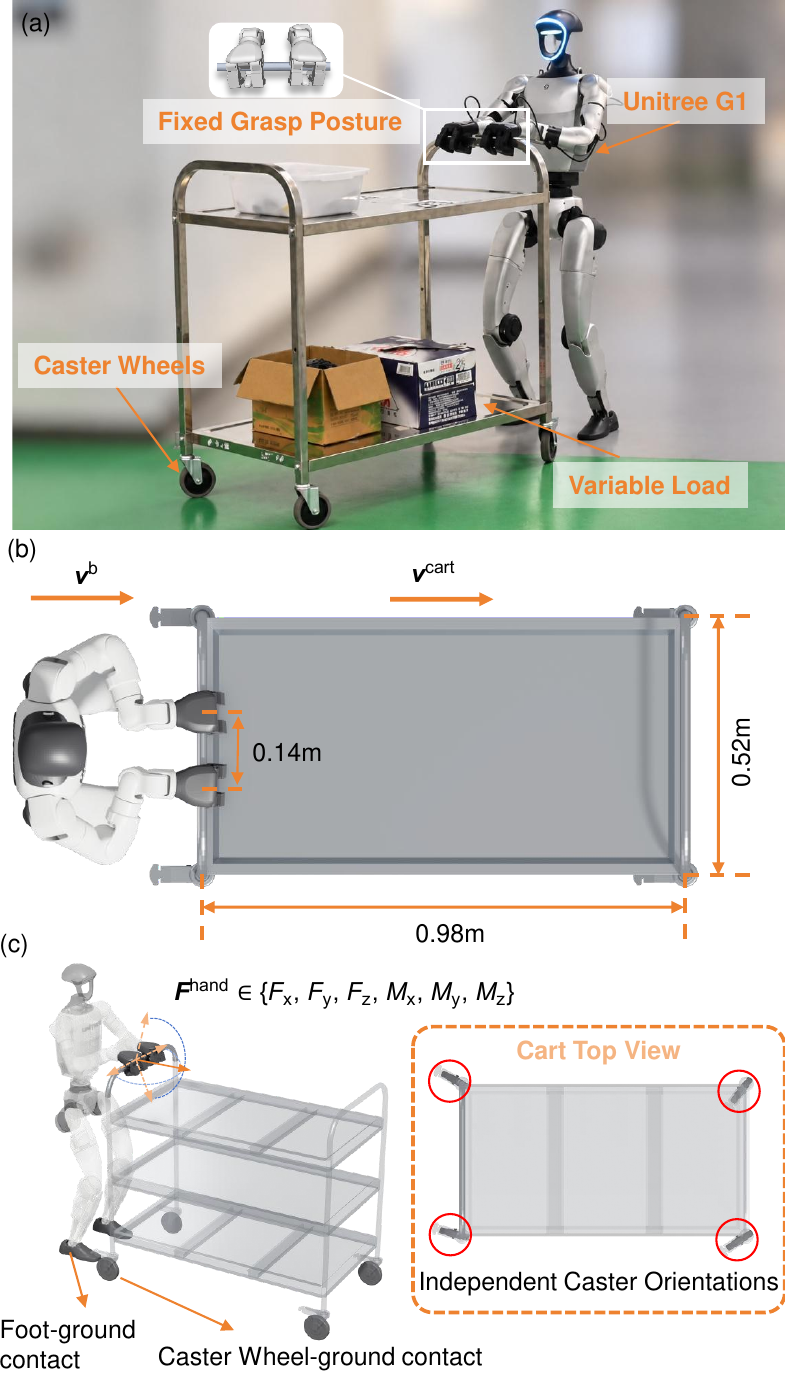}
\caption{(a) A Unitree G1 grasps the cart handle with a fixed grasp posture, while the cart rolls on caster wheels and carries a variable payload.
(b) Top view of the robot--cart system, indicating the bimanual grasp spacing, the cart dimensions, the robot base velocity $\basevel$, and the cart velocity $\carttwist$.
(c) Contact interactions that govern the coupled robot--cart dynamics.
}
\label{fig:problem_setup}
\end{figure}

A humanoid robot grasps the cart handle with both hands and drives the unactuated cart through its walking and hand forces, as illustrated in \Cref{fig:problem_setup}.
The robot must regulate the planar cart velocity $\carttwist_{\tidx}$ to track a commanded velocity $\cartcmd_{\tidx}$ while maintaining its balance and its grasp throughout the motion.

We split the state of the robot--cart system at control step $\tidx$ into an observable state $\obsstate_{\tidx}$ and an unobservable state $\unobsstate_{\tidx}$, so that the system $\sysmodel$ evolves as
\begin{equation}
\sysstate_{\tidx+1} = \sysmodel\big(\obsstate_{\tidx},\, \unobsstate_{\tidx},\, \wholeaction_{\tidx}\big),
\label{eq:full_model}
\end{equation}
where $\sysstate_{\tidx} = (\obsstate_{\tidx},\, \unobsstate_{\tidx})$ and $\wholeaction_{\tidx}$ is the whole-body action.
The observable state contains the signals available on the robot, namely proprioception, the relative cart heading $\cartyaw_{\tidx}$ from the head camera, and the velocity command $\cartcmd_{\tidx}$.
However, the hands carry no force sensors to measure the hand wrenches $\handwrench_{\tidx}$, everyday carts carry no motion sensors to measure the cart velocity $\carttwist_{\tidx}$, and the payload is unknown before the interaction begins, so the robot cannot directly perceive its interaction with the cart.
This interaction information is critical because the four swivel casters impose no fixed-wheel constraint, which leaves the lateral and yaw motions of the cart primarily driven by the hand wrenches, while changes in the payload mass and distribution between transport tasks alter the mapping from hand wrenches to cart motion.
Precise cart control thus depends on both $\obsstate_{\tidx}$ and $\unobsstate_{\tidx}$, and the latter must be inferred rather than measured.

In the system $\sysmodel$, the hand wrenches drive the cart while their reactions disturb the balance of the robot, so locomotion and manipulation are dynamically coupled.
Directly training a policy on the system $\sysmodel$ is difficult because a single policy must learn locomotion and manipulation at once, which leads to inefficient exploration~\citep{zhang2026falcon}, and it must do so under contact conditions that it cannot observe.
We therefore decompose the system into two subsystems, $\locomodel$ and $\pushmodel$, that are trained separately, while the physical system itself remains coupled.
Because the policies act on the observable state, we describe each subsystem by the evolution of this state.
From the perspective of locomotion, the cart affects the balance of the robot mainly through the reaction wrenches at the hands, which allows locomotion training to replace the cart with a hand wrench disturbance $\envwrench_{\tidx}$ representing these reactions.
This yields the first subsystem, in which the robot learns to walk stably under the reactions of cart pushing, as
\begin{equation}
\robotobs_{\tidx+1} = \locomodel\big(\robotobs_{\tidx},\, \envwrench_{\tidx},\, \wholeaction_{\tidx}\big),
\label{eq:model_loco}
\end{equation}
where $\robotobs_{\tidx}$ is the robot part of $\obsstate_{\tidx}$, and $\wholeaction_{\tidx}$ comprises the leg action $\legaction_{\tidx}$ of the locomotion policy $\policygmt$ and bounded upper-body posture offsets.
However, because $\locomodel$ contains no cart, it can stabilize the walking of the robot but cannot make the cart track the commanded velocity, which calls for a second subsystem with the cart attached.
In this subsystem, $\policygmt$ is frozen because it has already learned to keep balance, and the task policy $\policytask$ only learns to control the cart through residual actions $\fullaction_{\tidx}$ that act on top of $\policygmt$, without relearning walking.
Unlike in $\locomodel$, the hand wrenches in $\pushmodel$ are no longer applied disturbances but arise from the contact between the robot and the cart, and they depend on the payload and the cart motion.
Because these unobservable states are not available to the policy, $\pushmodel$ forms a partially observable Markov decision process.
Since they shape how proprioception evolves, they can be estimated from the proprioceptive history $\interactionhistory_{\tidx}$ through an estimator $\unobsest$, as
\begin{equation}
\unobshat_{\tidx} = \unobsest\big(\interactionhistory_{\tidx}\big).
\label{eq:unobs_estimate}
\end{equation}
With this estimate in place of the unobservable states, the second subsystem as seen by the task policy becomes
\begin{equation}
\obsstate_{\tidx+1} = \pushmodel\big(\obsstate_{\tidx},\, \unobshat_{\tidx},\, \fullaction_{\tidx}\big).
\label{eq:model_push}
\end{equation}
The task policy is then trained to maximize the expected discounted return
\begin{equation}
J(\policytask) = \mathbb{E}\!\left[ \sum_{\tidx=0}^{\horizon-1} \discount^{\tidx} \taskreward_{\tidx} \right],
\label{eq:rl_objective}
\end{equation}
where $\discount \in (0, 1)$ is the discount factor, $\horizon$ is the episode horizon, and $\taskreward_{\tidx}$ is the task reward defined in \Cref{eq:task_reward}.

\section{Method}
\label{sec:method}

For $\locomodel$, in which hand wrench disturbances stand in for the cart, we pre-train a locomotion prior under randomized hand wrenches and bounded upper-body posture offsets in \Cref{subsec:loco_prior}, so that the robot remains balanced under the reactions and postures that arise in cart pushing.
For $\pushmodel$, we realize the estimator $\unobsest$ in \Cref{subsec:latent_inference} with a force-guided interaction inference module, which infers force and state latents from the history $\interactionhistory_{\tidx}$ to represent the unobservable states.
Using these latents, the task policy in \Cref{subsec:whole_body_coordination} adds base-velocity command adjustments and arm, waist, and leg joint residuals to the frozen prior, coordinating the whole body to track the commanded cart velocity.

\begin{figure*}[!t]
\centering
\includegraphics[width=\textwidth]{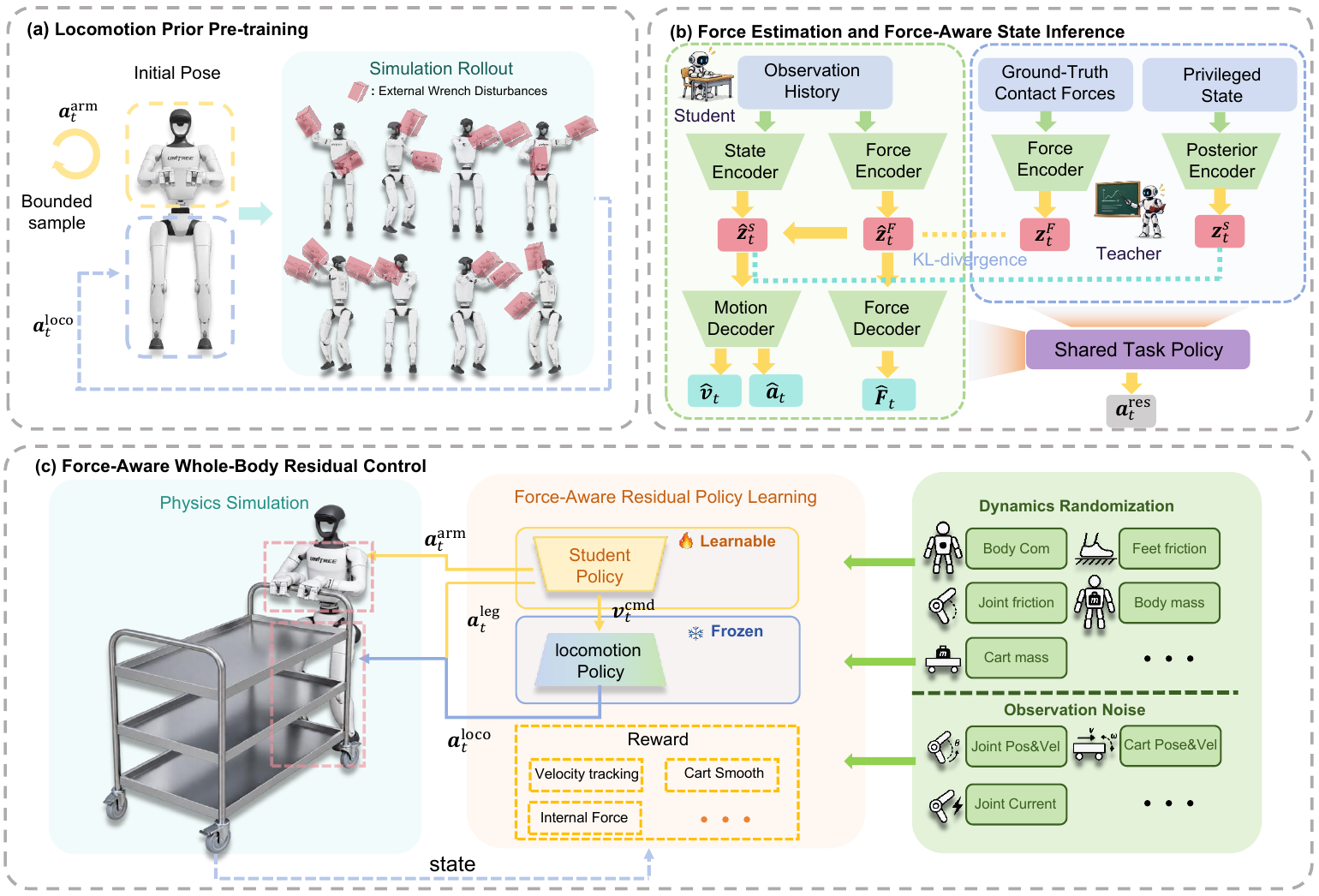}
\caption{(a) The locomotion prior is pre-trained to track base-velocity commands under random palm wrenches and bounded arm and waist posture variations.
(b) The teacher encodes privileged contact forces and states into force and state latents, whereas the student infers both latents from onboard observation histories and reconstructs the contact forces and the velocities and accelerations of the robot and the cart.
(c) Using the inferred latents, the task policy outputs base-velocity command adjustments for the frozen prior together with arm, waist, and leg joint residuals, and it is trained under dynamics randomization and observation noise with rewards for cart-velocity tracking, cart-motion smoothness, and internal-force suppression.
}
\label{fig:framework}
\end{figure*}

\subsection{Locomotion Prior for Interactive Cart Pushing}
\label{subsec:loco_prior}

The humanoid controls the cart's motion through coordinated walking and force application at the hands.
During this process, the robot adjusts its arm and waist configuration to regulate hand-contact wrenches while experiencing reaction forces and moments from the cart.
As formulated in \Cref{sec:problem}, $\locomodel$ requires the robot to maintain balance and track base-velocity commands under these combined effects.
We therefore pre-train the locomotion prior under simultaneous upper-body posture variations and hand loading to provide support for subsequent cart manipulation (\Cref{fig:framework}(a)).

During pre-training, the robot tracks base-velocity commands without the cart attached.
Three-axis external forces and moments are continuously applied at each palm, implementing the hand-wrench disturbances that replace the cart's reactions in the locomotion subsystem.
Meanwhile, the arm and waist joint targets vary around the nominal pushing posture within bounds matched to the upper-body residual range in the second stage.
These training conditions expose the locomotion prior to both the reaction loads and the configuration changes associated with cart manipulation, allowing locomotion support to be learned before introducing cart-motion control.

The locomotion prior is trained to track base-velocity commands while maintaining balance under the posture variations and hand loading described above.
Its reward combines four groups of weighted terms as
\begin{equation}
\rewloco_{\tidx}
=
\locotrack_{\tidx}
+
\locobalance_{\tidx}
+
\locogait_{\tidx}
-
\locoregcost_{\tidx}.
\label{eq:loco_reward}
\end{equation}
The principal velocity-tracking term, which rewards accurate forward and lateral base velocities and yaw rate, is
\begin{equation}
\locotrack_{\tidx}
=
\sum_{i\in\{x,y,\omega\}}
\rewweight{i}
\exp\!\left[
-\frac{\bigl([\basevel_{\tidx}]_i-[\cmdvel_{\tidx}]_i\bigr)^2}{\locotrackwidth_i}
\right],
\label{eq:loco_velocity_reward}
\end{equation}
where $\basevel_{\tidx}=[v_{x,\tidx}^{\mathrm b},v_{y,\tidx}^{\mathrm b},\omega_{z,\tidx}^{\mathrm b}]\transpose$ contains the measured base-motion components, $\cmdvel_{\tidx}$ denotes their commands, and $\rewweight{i}>0$ and $\locotrackwidth_i>0$ specify the tracking weight and error width, respectively.

The balance terms regulate base height and penalize excessive tilt and undesired base motion.
The gait terms regulate foot contact and swing behavior, including clearance and slip.
The nonnegative regularization cost penalizes abrupt action changes, excessive joint effort, and joint-limit violations.
Together, these terms encourage the robot to maintain commanded motion under loading while regulating its supporting posture, foot behavior, and actuation.
Episodes terminate upon excessive body tilt or designated non-foot contacts.

After pre-training, the parameters of the locomotion prior are frozen.
In $\pushmodel$, the task policy uses this prior to coordinate base-velocity commands and bounded joint residuals, learning cart-motion regulation on top of the acquired locomotion support.

\subsection{Latent Interaction Dynamics Inference under Partial Observability}
\label{subsec:latent_inference}
The humanoid transmits forces to the cart through its hand--handle contacts, driving the cart's translation and rotation.
Together with the cart's inertia and wheel--ground interactions, these contact forces determine its acceleration and thereby influence the evolution of its velocity.
Meanwhile, hand loading is reflected in the robot's joint torques and joint motion, while the history of the relative cart heading provides partial information about the motion response.
These onboard signals contain information linking contact loading to the resulting motion, providing a basis for inferring interaction states without external force sensors.

Building on this relationship, we construct a teacher--student inference architecture that uses contact-force information for both whole-body coordination and motion-state estimation (\Cref{fig:framework}(b)).
During training, the teacher encodes the unobservable states into a force latent and a state latent from privileged simulation information, and the task policy learns to act on these latents.
This privileged information is unavailable at deployment, so the student predicts both latents from the observation history $\interactionhistory_{\tidx}$, thereby recovering the unobservable states from signals available on the robot.
The student thus realizes the estimator $\unobsest$ of \Cref{eq:unobs_estimate}, and the predicted latents form the estimate $\unobshat_{\tidx}=(\studentforce_{\tidx},\,\studentstate_{\tidx})$.
The student first infers a force latent from observation histories containing joint motion, joint torques, and the relative cart heading.
This latent is fed directly to the task policy to provide contact-loading information; it is also fused with the observation history to estimate the motion states of the robot and the cart and produce a state latent for the task policy.

During training, the teacher constructs separate force and state latents from privileged simulation information.
The force encoder maps the reference contact forces $\handwrenchlabel_{\tidx}$ to a force latent $\teacherforce_{\tidx}$, while the state encoder maps true velocities and other privileged information $\privstate_{\tidx}$ to a state latent $\teacherstate_{\tidx}$.
Explicit contact-force channels are excluded from $\privstate_{\tidx}$.
Both latents enter the task policy and provide supervision targets for the student, as
\begin{equation}
\teacherforce_{\tidx}=\teacherforceenc(\handwrenchlabel_{\tidx}),
\qquad
\teacherstate_{\tidx}=\teacherstateenc(\privstate_{\tidx}).
\label{eq:teacher_interaction_latents}
\end{equation}

To recover these representations from onboard signals, the student first processes the observation history $\interactionhistory_{\tidx}$ through a causal temporal encoder, producing a sequence of force latents $\forcelatentseq_{\tidx}$.
The current force latent $\studentforce_{\tidx}$ is aligned with the teacher's $\teacherforce_{\tidx}$ and decoded through a shared decoder $\forcedecoder$ to reconstruct the reference contact forces.
These two objectives train the student to recover the teacher's force representation while retaining contact-loading information.
The resulting force latent is supplied directly to the task policy and also used in subsequent motion-state inference.

To relate contact loading to the motion response, the student fuses the force-latent sequence with the corresponding observation frames.
A temporal convolutional branch processes the full 32-frame history, while a second branch processes the most recent six frames to retain recent observation details.
The fused features $\interactionfeatures_{\tidx}$ produce the state latent and auxiliary velocity and acceleration estimates, as
\begin{equation}
\forcelatentseq_{\tidx}=\studentforceenc(\interactionhistory_{\tidx}),\quad
\handwrenchest_{\tidx}=\forcedecoder(\studentforce_{\tidx}),\quad
\interactionfeatures_{\tidx}=\interactionfusion(\interactionhistory_{\tidx},\forcelatentseq_{\tidx}),
\label{eq:student_force_features}
\end{equation}
\begin{equation}
(\studentstate_{\tidx},\motionvelest_{\tidx},\motionaccest_{\tidx})=\statehead(\interactionfeatures_{\tidx}),
\label{eq:student_interaction_inference}
\end{equation}
where $\statehead$ is the state decoder, whose output is partitioned into the estimated state latent $\studentstate_{\tidx}$, the velocity estimate $\motionvelest_{\tidx}$, which stacks the estimated planar linear velocity and yaw rate of the cart and the planar base linear velocity of the robot, and the acceleration estimate $\motionaccest_{\tidx}$, which contains the corresponding accelerations.

The state latent $\studentstate_{\tidx}$ is aligned with the teacher's $\teacherstate_{\tidx}$, allowing it to replace the privileged representation used by the task policy during deployment.
In parallel, the velocity and acceleration estimates are supervised by simulation reference values, requiring the shared features to explain explicit motion states.
Latent alignment therefore trains the student to recover the policy's interaction representation, while motion supervision constrains the information retained in the shared features.
Because these features depend on the force-latent sequence, motion-estimation errors also propagate through the force encoder.
The force representation is thus trained both to reconstruct contact loading and to support motion-state inference.
The task policy receives the force and state latents, while the velocity and acceleration estimates serve as auxiliary supervised outputs.

The student is optimized with the joint objective
\begin{equation}
\lossstudent
=\lossalign+\estlossweight{F}\lossforce
+\estlossweight{v}\lossmotionvel+\estlossweight{a}\lossmotionacc,
\label{eq:student_joint_loss}
\end{equation}
\begin{equation}
\lossalign
=\left\|\studentforce_{\tidx}-\stopgrad(\teacherforce_{\tidx})\right\|_2
+\left\|\studentstate_{\tidx}-\stopgrad(\teacherstate_{\tidx})\right\|_2.
\label{eq:student_latent_alignment}
\end{equation}
Here, $\stopgrad$ denotes the stop-gradient operator, keeping the teacher targets fixed during each student update.
The force, velocity, and acceleration losses are mean-squared errors between normalized estimates and their simulation references, and $\estlossweight{F},\estlossweight{v},\estlossweight{a}$ are positive weights.
During teacher optimization, force reconstruction and variance regularization constrain the force latent, while a regularization term draws the state latent toward the student estimate to encourage recoverability from onboard observations.
Training periodically uses student-estimated latents to collect rollouts, exposing the inference network to states encountered under its own estimation errors.

During training, the direct upper-body joint-torque channels in the task-policy observations are progressively suppressed, while these measurements remain available to the student inference network.
Contact-force information therefore reaches the task policy through the force latent rather than through the raw torques, which makes the policy explicitly force-aware.
At deployment, the privileged encoders are removed, and the shared task policy uses recent observations together with the estimated force and state latents for whole-body residual coordination.

\subsection{Force-Aware Whole-Body Residual Coordination}
\label{subsec:whole_body_coordination}

The task policy uses recent onboard observations and the force and state latents inferred in \Cref{subsec:latent_inference} to coordinate the humanoid's motion for cart-velocity tracking (\Cref{fig:framework}(c)).
The locomotion prior provides the underlying walking response, with its parameters frozen during task training.
The task policy learns to adjust this response and the upper-body configuration according to the inferred interaction, so that locomotion and manipulation jointly serve the commanded cart motion.
Whereas $\policygmt$ tracks robot base-velocity commands in $\locomodel$, $\policytask$ is optimized in $\pushmodel$ for the cart's forward and lateral velocities and yaw rate.

The task policy maps the recent observations and the inferred latents to the residual action
\begin{equation}
\fullaction_{\tidx} = \policytask\big(\observation_{\tidx},\, \studentforce_{\tidx},\, \studentstate_{\tidx}\big), \quad \fullaction_{\tidx} = \big[\cmdres_{\tidx},\, \resaction_{\tidx},\, \legresout_{\tidx}\big].
\label{eq:task_policy}
\end{equation}
The policy produces 30 coordinated action components, namely three base-velocity command adjustments, 15 arm and waist joint residuals, and 12 leg joint residuals.
The base-command adjustments modify the motion requested from the locomotion prior, allowing the robot's walking response to be adapted to the desired cart motion.
The arm and waist residuals adjust the nominal pushing posture and thereby regulate hand-contact forces through joint control while maintaining the grasp.
The leg residuals modify the prior's leg targets to accommodate task-specific support adjustments under contact loading.
All three outputs are generated jointly, allowing base motion, force application, and leg support to be coordinated toward the same cart-motion objective.

The task reward encourages accurate and smooth cart motion while maintaining robot balance and hand contact, and it explicitly penalizes the internal load between the two hands.
Its weighted terms are grouped as
\begin{equation}
\taskreward_{\tidx}=\carttrack_{\tidx}-\tasksmoothcost_{\tidx}-\rewweight{\mathrm{int}}\costint_{\tidx}-\graspcost_{\tidx}+\tasksupport_{\tidx}-\locoregcost_{\tidx}.
\label{eq:task_reward}
\end{equation}
The tracking reward encourages the cart's forward and lateral velocities and yaw rate to follow their references, and the smoothness costs suppress cart-velocity fluctuations and abrupt changes in contact forces.
Because both hands hold the same handle, only part of the bimanual hand wrench moves the cart.
We separate this driving part through $\drivechannel_{\tidx}=\drivemap\handwrench_{\tidx}$, where the constant matrix $\drivemap\in\real^{4\times12}$ maps the bimanual hand wrench to the push force of each hand, the sum of the lateral forces, and the sum of the yaw moments.
The remaining internal wrench produces no cart motion and only consumes joint torque margin,
\begin{equation}
\intwrench_{\tidx}=\handwrench_{\tidx}-\capweight\drivemap\transpose\big(\drivemap\capweight\drivemap\transpose\big)^{-1}\drivemap\handwrench_{\tidx},
\label{eq:internal_wrench}
\end{equation}
and the internal-load cost penalizes it in proportion to the load capacity of each grasp direction,
\begin{equation}
\costint_{\tidx}=\big\|\capweight^{-1/2}\intwrench_{\tidx}\big\|_2^2,
\label{eq:internal_cost}
\end{equation}
where $\capweight$ is a diagonal matrix of squared grasp capacities.
The hand wrench is low-pass filtered before this decomposition, and the vertical load needed for frictional grip is not counted as internal.
The grasp cost penalizes grasp slip, abrupt force changes, and wrist saturation, while the support and regularization terms promote balanced locomotion and smooth, moderate control actions.

\begin{figure*}[!t]
\centering
\includegraphics[width=0.99\textwidth]{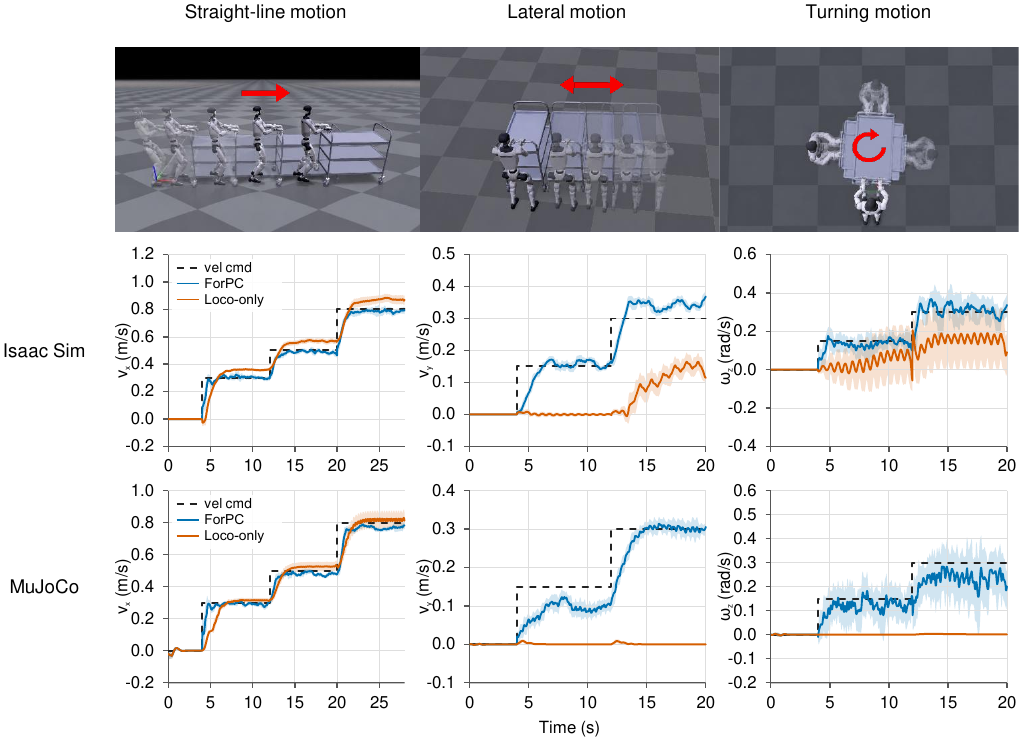}
\caption{Cart-velocity tracking by ForPC and Loco-only in Isaac Sim and MuJoCo.
Loco-only uses the Stage~I policy without task-level corrections.
Both policies are trained only in Isaac Sim, so the MuJoCo results serve as a sim-to-sim test; shaded bands denote standard deviations.
}
\label{fig:motion_response}
\end{figure*}

\section{Experiments}
\label{sec:experiments}
We evaluate the cart-pushing performance of ForPC in simulation and on a physical Unitree G1 from three aspects.
First, we assess whether ForPC tracks cart-velocity commands accurately with low velocity fluctuations in forward, lateral, and turning motions and maintains this accuracy over complete routes under different payloads (\Cref{subsec:motion_results}).
Second, we analyze whether the inferred latents retain contact-loading and cart-motion information and whether force awareness directs the bimanual forces toward driving the cart rather than toward mutual opposition between the hands (\Cref{subsec:interaction_results}).
Third, we test whether the simulation-trained policy transfers to the physical robot and accomplishes real-world cart transport without real-world training or fine-tuning (\Cref{subsec:sim2real}).

\subsection{Experimental Setup}
\label{subsec:setup}
The robot is a Unitree G1 humanoid with 12 leg joints, one waist yaw joint, and 14 arm joints, and the cart is a four-caster platform trolley with an empty mass of \SI{12}{kg}.
The robot carries three-fingered Dex3-1 hands held at a fixed grasp configuration, the waist roll and pitch joints are held at zero, and a RealSense camera in the head detects an AprilTag fixed to the cart to measure the relative cart heading.
The payload is the mass that is added to the cart.
The policies are trained in Isaac Sim with the Isaac Lab framework~\citep{mittal2023orbit} and are evaluated with fixed weights in Isaac Sim and in MuJoCo~\citep{todorov2012mujoco}.
The MuJoCo evaluation is therefore a sim-to-sim test in which the contact model, the solver, and the actuator integration differ from those of training.
Both the policy and the simulators run at \SI{50}{Hz} control rate, and the cart velocities are sampled at \SI{50}{Hz}.

\begin{table*}[!t]
\centering\small
\setlength{\tabcolsep}{3pt}
\renewcommand{\arraystretch}{1.12}
\caption{Cart-velocity tracking RMSE and RMSF in Isaac Sim and MuJoCo at two payloads.
Units are m/s for forward and lateral motion and rad/s for turning; $\downarrow$ indicates that lower is better, and bold marks the best value per simulator and payload.}
\label{tab:motion_results}
\begin{tabular*}{\textwidth}{@{\extracolsep{\fill}}lllcccccc@{}}
\toprule
\multirow{2}{*}{Simulator} & \multirow{2}{*}{\shortstack{Payload\\(kg)}} & \multirow{2}{*}{Method} & \multicolumn{2}{c}{Forward} & \multicolumn{2}{c}{Lateral} & \multicolumn{2}{c}{Turning} \\
\cmidrule(lr){4-5}\cmidrule(lr){6-7}\cmidrule(lr){8-9}
 & & & RMSE $\downarrow$ & RMSF $\downarrow$ & RMSE $\downarrow$ & RMSF $\downarrow$ & RMSE $\downarrow$ & RMSF $\downarrow$ \\
\midrule
\multirow{8}{*}{Isaac Sim} & \multirow{4}{*}{15} & Loco-only & 0.1768 & 0.1487 & 0.1564 & 0.1387 & 0.2989 & 0.2611 \\
 &  & Standard prior w Force-aware & 0.0816 & 0.0612 & 0.0908 & 0.0613 & 0.1067 & \textbf{0.1103} \\
 &  & Standard prior w/o Force-aware & 0.0984 & 0.0758 & 0.1421 & 0.0612 & 0.1311 & 0.1221 \\
 &  & \textbf{ForPC} & \textbf{0.0801} & \textbf{0.0590} & \textbf{0.0892} & \textbf{0.0591} & \textbf{0.1010} & 0.1203 \\
\cmidrule(l){2-9}
 & \multirow{4}{*}{25} & Loco-only & 0.2003 & 0.1449 & 0.1612 & 0.1320 & 0.2958 & 0.2453 \\
 &  & Standard prior w Force-aware & 0.0856 & 0.0546 & 0.1064 & 0.0523 & 0.1048 & 0.1031 \\
 &  & Standard prior w/o Force-aware & 0.1026 & 0.0680 & 0.1548 & 0.0516 & 0.1305 & 0.1061 \\
 &  & \textbf{ForPC} & \textbf{0.0851} & \textbf{0.0501} & \textbf{0.1022} & \textbf{0.0515} & \textbf{0.0941} & \textbf{0.0959} \\
\midrule
\multirow{8}{*}{MuJoCo} & \multirow{4}{*}{15} & Loco-only & 0.1393 & 0.1417 & 0.1494 & 0.1281 & 0.2830 & 0.2162 \\
 &  & Standard prior w Force-aware & 0.0736 & 0.0606 & 0.0799 & 0.0691 & 0.1149 & \textbf{0.1059} \\
 &  & Standard prior w/o Force-aware & 0.0882 & 0.0683 & 0.1159 & 0.0573 & 0.1286 & 0.1240 \\
 &  & \textbf{ForPC} & \textbf{0.0707} & \textbf{0.0573} & \textbf{0.0739} & \textbf{0.0551} & \textbf{0.1087} & 0.1109 \\
\cmidrule(l){2-9}
 & \multirow{4}{*}{25} & Loco-only & 0.1578 & 0.1383 & 0.1541 & 0.1206 & 0.2805 & 0.2046 \\
 &  & Standard prior w Force-aware & 0.0773 & 0.0539 & 0.0964 & 0.0526 & 0.1164 & \textbf{0.0958} \\
 &  & Standard prior w/o Force-aware & 0.0928 & 0.0632 & 0.1246 & 0.0596 & 0.1365 & 0.1074 \\
 &  & \textbf{ForPC} & \textbf{0.0731} & \textbf{0.0498} & \textbf{0.0834} & \textbf{0.0502} & \textbf{0.1032} & 0.0986 \\
\bottomrule
\end{tabular*}
\end{table*}

\begin{table*}[!t]
\centering\small
\setlength{\tabcolsep}{4pt}
\renewcommand{\arraystretch}{1.12}
\caption{Route-tracking results of four configurations at two payloads under matched speed limits.
$\downarrow$ and $\uparrow$ indicate that lower and higher are better, respectively; bold marks the best value per payload.}
\label{tab:route_results}
\begin{tabular*}{\textwidth}{@{\extracolsep{\fill}}llcccc@{}}
\toprule
\multirow{2}{*}{\shortstack{Payload\\(kg)}} & \multirow{2}{*}{Method} & \multicolumn{2}{c}{RMSE $\downarrow$} & \multirow{2}{*}{Completion time (s) $\downarrow$} & \multirow{2}{*}{Success rate (\%) $\uparrow$} \\
\cmidrule(lr){3-4}
 & & Cross-track (m) & Heading (deg) & & \\
\midrule
\multirow{4}{*}{15} & Loco-only & 0.2153 & 23.83 & 62.00 & 89.1 \\
 & Standard prior w Force-aware & 0.0658 & 5.70 & 42.49 & 93.8 \\
 & Standard prior w/o Force-aware & 0.1494 & 14.22 & 45.81 & 90.6 \\
 & \textbf{ForPC} & \textbf{0.0648} & \textbf{5.63} & \textbf{41.49} & \textbf{100} \\
\midrule
\multirow{4}{*}{25} & Loco-only & 0.2652 & 25.70 & 71.07 & 76.6 \\
 & Standard prior w Force-aware & 0.0735 & 6.36 & 43.47 & 95.3 \\
 & Standard prior w/o Force-aware & 0.1641 & 16.55 & 47.98 & 89.1 \\
 & \textbf{ForPC} & \textbf{0.0685} & \textbf{5.98} & \textbf{42.45} & \textbf{100} \\
\bottomrule
\end{tabular*}
\end{table*}

We compare ForPC with three configurations to evaluate the roles of task-level coordination, locomotion-prior training, and force awareness.
The first, locomotion-only, uses the Stage~I policy alone, taking the commanded cart velocity as its base-velocity command and maintaining the nominal bimanual contact posture without task-level corrections.
The second retains force-aware whole-body residual coordination but uses a locomotion prior trained without hand-wrench disturbances or upper-body posture randomization.
The third uses the same prior-training conditions as the second but additionally removes explicit force awareness from the residual coordination framework.
Full ForPC combines the locomotion prior trained under hand loading and upper-body posture variations with force-aware coordination of base-velocity command adjustments and arm, waist, and leg joint residuals.
All configurations are evaluated using the same commands, cart, and environmental conditions.

\subsection{Humanoid Cart-Pushing Control}
\label{subsec:motion_results}

We evaluate humanoid cart-pushing control through velocity tracking and complete-route tracking.

\subsubsection{Cart-Velocity Tracking}
\label{subsubsec:velocity_tracking}
To evaluate tracking accuracy and velocity fluctuations during humanoid cart pushing, we compare full ForPC with the three configurations introduced in \Cref{subsec:setup} through forward, lateral, and turning velocity-tracking tests at payloads of \SI{15}{kg} and \SI{25}{kg}.
At each payload, all configurations use identical velocity commands and resistance settings.
We apply sequences of constant commands separately for forward translation ($0.3$, $0.5$, and \SI{0.8}{m/s}), lateral translation ($0.15$ and \SI{0.3}{m/s}), and rotation about the vertical axis ($0.15$ and \SI{0.3}{rad/s}), holding each command for \SI{8}{s}.
Tracking errors and fluctuations in the cart's forward velocity, lateral velocity, and yaw rate are compared to assess the control performance of each configuration and its variation with payload.

\noindent\textbf{Evaluation metrics.}
\label{subsubsec:motion_metrics}
We use the root-mean-square error (RMSE) to evaluate velocity-tracking accuracy and RMS velocity fluctuation (RMSF) to quantify velocity variations within constant-command plateaus.
Let $\horizon$ denote the number of samples used to compute the metrics.
For each evaluated component, let $\metricvelocity_{\tidx}$ and $\metriccommand_{\tidx}$ denote the actual and commanded velocities in the same coordinate frame, and let $\metricerror_{\tidx}=\metricvelocity_{\tidx}-\metriccommand_{\tidx}$ denote their difference; actual velocities are unfiltered, and uncommanded components have a zero reference.
The following scalar formulas are applied separately to forward velocity, lateral velocity, and yaw rate.

\noindent\textbf{(1) RMSE.}
For a scalar error $\metricerror_{\tidx}$ over $\horizon$ samples, the RMSE is
\begin{equation}
\motionrmse=\sqrt{\frac{1}{\horizon}\sum_{\tidx=1}^{\horizon}\metricerror_{\tidx}^{2}},
\label{eq:rmse}
\end{equation}
which we use for all tracking and estimation errors in this section; for velocity tracking, $\metricerror_{\tidx}$ is the velocity error defined above.

\noindent\textbf{(2) RMSF.}
RMSF measures the root-mean-square deviation of velocity from its own command-plateau mean, excluding the plateau mean tracking error and changes between plateau means.
With the plateau $k(\tidx)$ that contains sample $\tidx$ and the mean velocity $\metricmean{k}$ over the evaluated samples of plateau $k$, it is
\begin{equation}
\motionripple=\sqrt{\frac{1}{\horizon}\sum_{\tidx=1}^{\horizon}\bigl(\metricvelocity_{\tidx}-\metricmean{k(\tidx)}\bigr)^2}.
\label{eq:motion_ripple}
\end{equation}
For every plateau, the squared RMSE equals the squared RMSF plus the squared mean error, and this metric includes the slow drift within a plateau because no frequency separation is applied.
All configurations use identical plateaus, and lower values indicate smaller tracking errors or weaker fluctuations.

\noindent\textbf{Results.}

\Cref{fig:motion_response} shows that ForPC responds to forward, lateral, and yaw commands in both simulators.
Once the forward motion is established, Loco-only follows the straight-line speed changes, but its lateral and turning responses remain weak in Isaac Sim and are nearly absent in MuJoCo.
These responses show that locomotion capability alone is insufficient for effective lateral and turning control of the dynamically coupled cart.
These maneuvers require precise regulation of the hand-contact forces and moments to drive the cart against the resistance while the robot maintains its balance under the resulting reactions.
By coordinating this interaction with whole-body support, ForPC follows the lateral and yaw commands more closely and extends the effective cart control beyond forward pushing.

\Cref{tab:motion_results} shows that force-aware whole-body coordination enables precise and smooth cart control that locomotion alone cannot provide.
ForPC attains the lowest RMSE in all twelve combinations of simulator, payload, and motion, reducing RMSE by up to 68.2\% and RMSF by up to 65.4\% relative to Loco-only.
The gap is largest in turning, where the cart must be rotated by a bimanual yaw moment that walking alone cannot generate.

Force awareness allows the policy to apply the hand forces that the cart actually requires.
With the same standard prior, adding force awareness lowers RMSE by up to 36.1\%, most strongly in lateral motion, which is driven mainly by lateral hand forces.
Perceiving the contact loading lets the policy correct the steady tracking offset caused by unmeasured resistance.

Training the locomotion prior under hand loading and posture variations further lowers RMSE by up to 13.5\%.
Because this prior already keeps the robot balanced under hand reactions, the residual policy can concentrate on cart control, and the benefit grows at \SI{25}{kg} and in MuJoCo, where the reactions are larger and the dynamics differ from training.
Overall, these results demonstrate that ForPC achieves accurate cart-velocity tracking under different payloads.

\subsubsection{Route Tracking}
\label{subsec:route_results}
\begin{figure}[!ht]
\centering
\includegraphics[width=\linewidth]{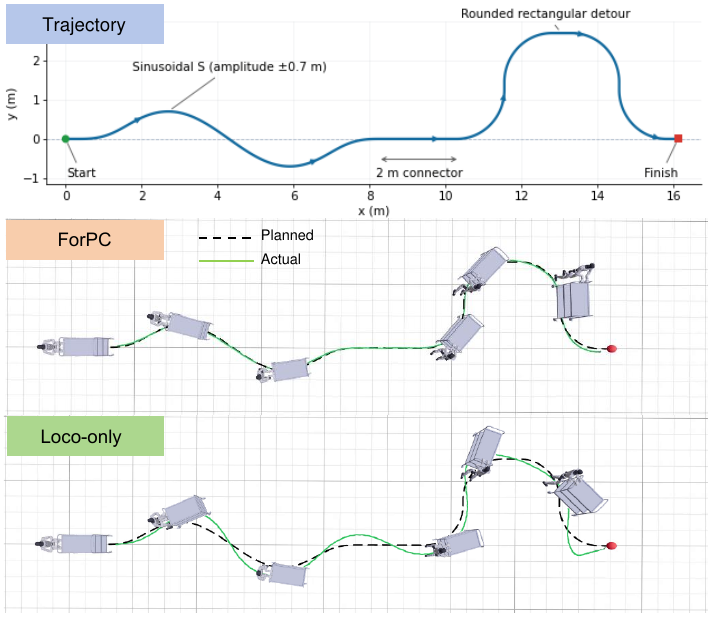}
\caption{Route tracking by ForPC and Loco-only in simulation.
Cart trajectories are overlaid on robot--cart snapshots.}
\label{fig:route_tracking}
\end{figure}

To evaluate how velocity control supports continuous cart transport, we test route tracking through changes in curvature and a final stop.
The route comprises an S-bend, a straight connector, and a rounded rectangular detour, ending at a target stopping pose (\Cref{fig:route_tracking}).
We compare full ForPC with the three configurations in \Cref{subsec:setup} at payloads of \SI{15}{kg} and \SI{25}{kg}, using cross-track RMSE, heading RMSE, completion time, and success rate.
Completion requires reaching the target within the time limit and the prescribed pose tolerances while the robot maintains its balance and the hand contact.
All configurations use the same reference route, route tracker, and speed limits.

\noindent\textbf{Evaluation metrics.}
We evaluate route tracking using cross-track RMSE, heading RMSE, completion time, and success rate, with $\horizon$ denoting the number of samples in the route-tracking interval.

\noindent\textbf{(1) Cross-track RMSE.}
The cross-track RMSE, reported in metres, applies \Cref{eq:rmse} to the signed lateral distance $\errlat_{\tidx}$ from the cart to the reference route at sample $\tidx$.

\noindent\textbf{(2) Heading RMSE.}
The heading RMSE applies \Cref{eq:rmse} to the heading error $\wrapangle(\worldyaw_{\tidx}-\pathheading_{\tidx})$, where $\worldyaw_{\tidx}$ is the cart heading in the world frame, $\pathheading_{\tidx}$ is the reference route heading at the nearest route point in the same frame, and $\wrapangle(\cdot)$ maps each angular difference to $(-\pi,\pi]$.
The error is computed in radians, and the resulting RMSE is converted to degrees for reporting.

\noindent\textbf{(3) Completion time.}
Completion time is the elapsed time, in seconds, from the start of route tracking until the cart first satisfies the endpoint completion criterion defined above.
Lower cross-track and heading RMSE indicate more accurate position and orientation tracking, respectively, whereas a shorter completion time indicates faster route completion under the matched speed limits.

\noindent\textbf{(4) Success rate.}
The success rate is the percentage of trials that satisfy the completion criterion defined above, computed over 640 trials for each configuration and payload, so a trial fails if the robot loses its balance or the hand contact, or does not reach the target pose within the time limit.

\noindent\textbf{Results.}
\Cref{fig:route_tracking} shows that ForPC follows the planned bends more closely, whereas Loco-only deviates from the route, particularly around the rounded rectangular detour.
This comparison illustrates how the ability to control lateral and turning motion supports tracking through changes in path curvature.

\Cref{tab:route_results} shows that the velocity-tracking advantage of ForPC carries over to complete routes.
ForPC achieves the best value in every metric at both payloads; relative to Loco-only, it reduces the cross-track RMSE by up to 74.2\% and the heading RMSE by up to 76.7\%, shortens the completion time by up to 40.3\%, and succeeds in every trial.
Accurate lateral and yaw control keeps the cart close to the route through successive bends, so the robot spends less time correcting deviations and completes the route sooner.

Force awareness remains the main contributor.
With the same standard prior, it reduces the cross-track RMSE by up to 56.0\% and the heading RMSE by up to 61.6\%, showing that the precise cart control enabled by force awareness allows ForPC to handle the lateral and heading errors that arise along curved routes.

Training the locomotion prior under hand loading further lowers the cross-track RMSE by up to 6.8\% and raises the success rate from 93.8\% and 95.3\% to 100\%, indicating that its balance support keeps the robot from failing over long transports.

As the payload increases from \SI{15}{kg} to \SI{25}{kg}, the cross-track RMSE of Loco-only rises by 23.2\% and its success rate drops from 89.1\% to 76.6\%, whereas the cross-track RMSE of ForPC changes only slightly and its success rate remains 100\%.
These results demonstrate that ForPC achieves accurate and reliable route transport under different payloads.

\subsection{Force-Latent Analysis and Bimanual Coordination}
\label{subsec:interaction_results}
\Cref{subsec:motion_results} shows that ForPC tracks cart velocities and routes more precisely than the compared configurations.
This subsection examines how fusing the force latent contributes to this precision.
We first decode the inferred latents into hand forces and cart velocities to test whether they retain contact-loading and cart-motion information, and then examine whether force awareness directs the bimanual forces toward driving the cart rather than toward mutual opposition between the hands.

\begin{table}[!ht]
\centering\footnotesize
\setlength{\tabcolsep}{3pt}
\renewcommand{\arraystretch}{1.1}
\caption{ForPC estimation accuracy in Isaac Sim, with force components expressed in the cart-heading frame.}
\label{tab:interaction_estimation}
\begin{tabular*}{\linewidth}{@{\extracolsep{\fill}}lcc@{}}
\toprule
Quantity & RMSE (N) $\downarrow$ & $R^2$ $\uparrow$ \\
\midrule
Left-hand forward force & 7.54 & 0.97 \\
Left-hand lateral force & 4.32 & 0.78 \\
Left-hand vertical force & 4.32 & 0.95 \\
Right-hand forward force & 7.36 & 0.97 \\
Right-hand lateral force & 4.28 & 0.78 \\
Right-hand vertical force & 4.28 & 0.95 \\
\midrule
Quantity & \multicolumn{2}{c}{RMSE (m/s, rad/s) $\downarrow$} \\
\midrule
Cart forward velocity & \multicolumn{2}{c}{0.035} \\
Cart lateral velocity & \multicolumn{2}{c}{0.035} \\
Cart yaw rate & \multicolumn{2}{c}{0.050} \\
\bottomrule
\end{tabular*}
\end{table}

\begin{table*}[!t]
\centering\small
\setlength{\tabcolsep}{4pt}
\renewcommand{\arraystretch}{1.12}
\caption{Internal-force ratio $\intratio$ in Isaac Sim; lower is better.}
\label{tab:internal_load_results}
\begin{tabular*}{\textwidth}{@{\extracolsep{\fill}}llccc@{}}
\toprule
Payload (kg) & Method & Forward (\%) & Lateral (\%) & Turning (\%) \\
\midrule
\multirow{4}{*}{15} & Loco-only & 83.6 & 83.7 & 78.2 \\
 & Standard prior w Force-aware & 84.7 & 61.5 & 57.3 \\
 & Standard prior w/o Force-aware & 96.9 & 85.3 & 89.2 \\
 & \textbf{ForPC} & \textbf{81.8} & \textbf{61.3} & \textbf{55.6} \\
\midrule
\multirow{4}{*}{25} & Loco-only & 81.7 & 81.5 & 75.2 \\
 & Standard prior w Force-aware & 78.7 & 57.0 & 53.3 \\
 & Standard prior w/o Force-aware & 96.1 & 82.9 & 87.9 \\
 & \textbf{ForPC} & \textbf{78.4} & \textbf{56.9} & \textbf{51.9} \\
\bottomrule
\end{tabular*}
\end{table*}

\subsubsection{Information Encoded in the Inferred Latents}
\label{subsec:estimator_results}
We test whether the student's latents retain contact-loading and cart-motion information during cart pushing in Isaac Sim, evaluating the force estimates over cart-pushing runs under various operating conditions and the cart-velocity estimates at a payload of \SI{25}{kg}.
The evaluated outputs are the three contact-force components of each hand and the cart's forward velocity, lateral velocity, and yaw rate.
Predictions and simulation references use matching coordinate frames, signs, and timestamps.
Hand forces are decoded from the force latent through $\forcedecoder$ in \Cref{eq:student_force_features}, and cart velocities are taken from the state decoder $\statehead$ in \Cref{eq:student_interaction_inference}, whose input features depend on the force-latent sequence.

\noindent\textbf{Evaluation metrics.}
For each force or velocity component, the estimation RMSE applies \Cref{eq:rmse} to the error $\estimatedsignal_{\tidx}-\referencesignal_{\tidx}$ between the predicted value $\estimatedsignal_{\tidx}$ and its simulation reference $\referencesignal_{\tidx}$ at sample $\tidx$.
This metric compares an estimate with its actual reference, whereas the velocity-tracking RMSE in \Cref{subsubsec:velocity_tracking} compares actual motion with the commanded motion.
For the force components, we also report the coefficient of determination
\begin{equation}
R^2=1-\frac{\sum_{\tidx=1}^{\horizon}\bigl(\referencesignal_{\tidx}-\estimatedsignal_{\tidx}\bigr)^2}{\sum_{\tidx=1}^{\horizon}\bigl(\referencesignal_{\tidx}-\referencemean\bigr)^2},
\label{eq:r2}
\end{equation}
where $\referencemean$ is the mean of the reference over the $\horizon$ samples; $R^2$ measures the fraction of the reference variation explained by the estimate, and values closer to one indicate closer agreement.

\noindent\textbf{Results.}
\Cref{tab:interaction_estimation} reports the decoding accuracy.
The component-wise force RMSE ranges from \SI{4.32}{N} to \SI{7.54}{N} for the left hand and from \SI{4.28}{N} to \SI{7.36}{N} for the right hand, with $R^2$ values of 0.97 for the forward forces, 0.95 for the vertical forces, and 0.78 for the lateral forces, indicating that the force latent encodes the contact loading of each hand.
At a payload of \SI{25}{kg}, the cart-velocity RMSE is \SI{0.035}{m/s} for both linear components and \SI{0.050}{rad/s} for the yaw rate, indicating that the fused features also capture the cart's motion response.

\subsubsection{Bimanual Force Utilization}
\label{subsec:force_results}
We compare the four configurations in \Cref{subsec:setup} in matched Isaac Sim pushing trials at payloads of \SI{15}{kg} and \SI{25}{kg}.
All configurations use the same cart-velocity command sequences, initial conditions, and evaluation duration for each motion type in \Cref{subsubsec:velocity_tracking}.
For every configuration, including those without a force estimator, the hand loads are taken from the simulation-reference hand wrenches.

\noindent\textbf{Evaluation metrics.}
Only the forward force, lateral force, and yaw moment of each hand act in the plane of cart motion.
The vertical forces and the remaining moments mainly sustain the grasp and are transmitted to the ground through the cart wheels; they are therefore excluded from this evaluation.
Let $\planarsel\in\real^{6\times12}$ select these six components from $\handwrench_{\tidx}$.
Because $\capweight$ is diagonal and $\drivemap$ acts only on these components, \Cref{eq:internal_wrench} splits the capacity-normalized horizontal load into two orthogonal parts: a driving part that moves the cart and an internal part that cancels between the hands.
The internal-force ratio over $\horizon$ samples is
\begin{equation}
\intratio=\frac{\sum_{\tidx=1}^{\horizon}\big\|\planarsel\capweight^{-1/2}\intwrench_{\tidx}\big\|_2^2}{\sum_{\tidx=1}^{\horizon}\big\|\planarsel\capweight^{-1/2}\handwrench_{\tidx}\big\|_2^2},
\label{eq:internal_force_ratio}
\end{equation}
where identical filtering and grasp capacities are applied to all configurations.
This ratio lies in $[0,1]$, and $1-\intratio$ is the fraction of the applied horizontal hand load that drives the cart.
A lower value indicates that less bimanual effort is lost to mutual opposition between the hands.
The ratio is computed separately for forward, lateral, and turning trials and is interpreted together with the tracking errors in \Cref{tab:motion_results} to verify that the cart remains accurately controlled.

\noindent\textbf{Results.}
\Cref{tab:internal_load_results} shows that force awareness determines whether the bimanual effort drives the cart or is lost to mutual opposition between the hands.
Without force awareness, the standard prior yields the highest internal-force ratio in every case, up to 14.4 percentage points above Loco-only, because its upper-body residuals act without perceiving the opposition they create.
Adding force awareness lowers the ratio by up to 34.6 percentage points and, in lateral and turning motion, increases the fraction of the horizontal hand load that drives the cart by up to 4.0 times.
The effort lost to opposition without force awareness is consistent with the larger lateral RMSE of this configuration in \Cref{tab:motion_results}.

ForPC remains within 3.0 percentage points of the standard prior with force awareness, indicating that the force distribution is set mainly by force awareness, whereas the trained prior contributes through the tracking accuracy and success rate reported in \Cref{subsec:motion_results}.

In forward motion, all configurations show high ratios with small differences.
A plausible explanation is that steady forward pushing requires only a small driving force against rolling resistance, so the inward grip forces that cancel between the hands dominate the horizontal load.
For every configuration, the ratio also decreases from \SI{15}{kg} to \SI{25}{kg}, because the heavier cart requires larger driving forces.

These results demonstrate that force awareness enables ForPC to direct its bimanual effort toward driving the cart.

\subsection{Real-World Deployment}
\label{subsec:sim2real}
We deploy ForPC on a physical Unitree G1 to examine whether the simulation-trained controller transfers to hardware.
The deployed networks, comprising the normalizers, the student inference network, and the task policy, are taken directly from simulation training without real-world training or fine-tuning.
On hardware, the policy relies only on proprioception, joint torques, and the relative cart heading obtained from the AprilTag; neither the hand forces nor the cart velocity is measured, and no payload information is provided.

\begin{figure}[!ht]
\centering
\includegraphics[width=\linewidth]{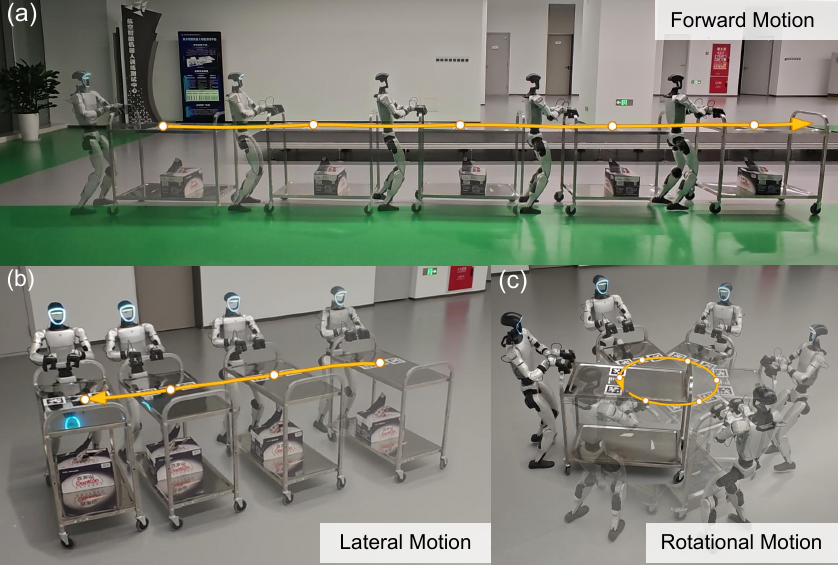}
\caption{(a) ForPC deployed on a physical Unitree G1 pushes the cart forward along a straight line.
(b) The deployed policy moves the cart laterally.
(c) The deployed policy rotates the cart.
}
\label{fig:real_motions}
\end{figure}

\begin{figure*}[!t]
\centering
\includegraphics[width=\textwidth]{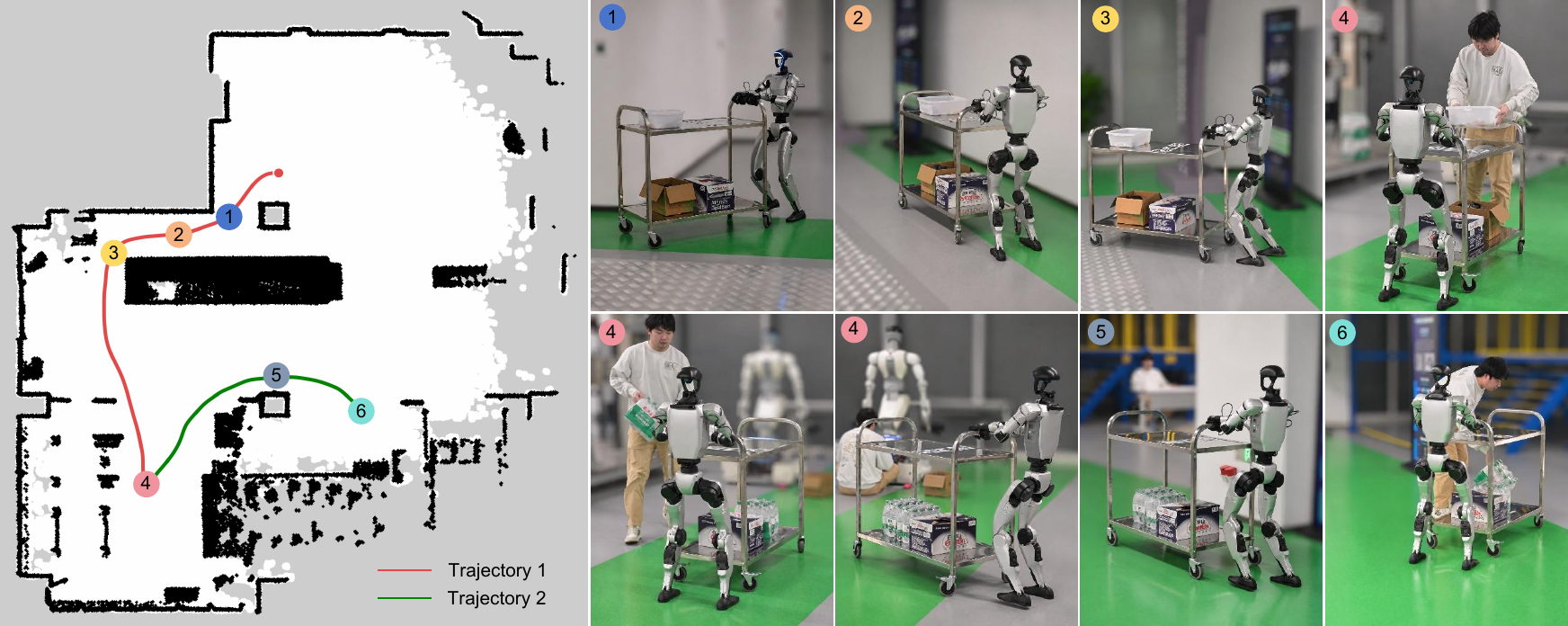}
\caption{A physical Unitree G1 running ForPC within a navigation system transports materials.
The map shows Trajectory 1 (red) and Trajectory 2 (green), and the numbered markers indicate where the corresponding snapshots were taken.
(1) The robot turns into a narrow passage.
(2) The robot moves straight through the narrow passage.
(3) The robot turns onto a straight segment.
(4) A worker removes materials from the cart, a worker places materials onto the cart, and the robot turns in place.
(5) Under the same policy, the robot pushes the cart with a different payload to complete the material transport.
(6) A worker removes the materials from the cart.
}
\label{fig:navigation_transport}
\end{figure*}

\Cref{fig:real_motions} shows three basic motions on a flat indoor floor.
In forward motion, the robot walks behind the cart and pushes it along an approximately straight line through a long corridor.
In lateral motion, the robot steps sideways with the cart, which translates across the floor while both hands remain on the handle.
In rotational motion, the robot walks around the cart while the cart turns about a nearly fixed point.
In lateral and rotational motion, the cart is driven mainly by lateral hand forces or a yaw moment rather than by forward pushing, while the legs reposition the body around the cart.
These motions therefore require the coordinated arm, waist, and leg adjustments described in \Cref{subsec:whole_body_coordination}.

We further integrate ForPC into a navigation system for material transport, a common logistics task, as shown in \Cref{fig:navigation_transport}.
The navigation system plans the route on a prebuilt map and issues cart-velocity commands, which ForPC executes through whole-body motion.
Along Trajectory 1, the robot pushes the cart into a narrow passage, moves straight through it, turns onto a long straight segment, and reaches the transfer station.
At this station, a worker unloads the transported materials and places a different payload on the cart, after which the robot turns the cart in place.
The robot then pushes the reloaded cart along Trajectory 2 to the destination, where the materials are unloaded.
The same policy executes both trajectories consecutively, without resetting the controller or adjusting any parameter after the payload change.

These experiments show that the simulation-trained policy transfers to the physical robot without fine-tuning, sustains coordinated pushing in forward, lateral, and rotational motion, and, in a real transport task with payload changes, pushes a cart carrying up to \SI{30}{kg} at \SI{0.5}{m/s}.

\section{Conclusion}
\label{sec:conclusion}
This paper presented ForPC, a force-aware whole-body control framework for precise humanoid cart pushing.
A locomotion prior trained under hand loading and upper-body posture variations provides balance support, a force-guided interaction inference module infers force and state latents from onboard signals without external force sensors, and a residual policy uses these latents to coordinate the arms, waist, and legs.
Sim-to-sim experiments show that ForPC reduces cart-velocity RMSE by up to 68.2\% and cross-track RMSE by up to 74.2\% relative to Loco-only, succeeds in every route trial, and directs more of the bimanual effort toward driving the cart.
Trained solely in simulation, the policy transfers to a physical Unitree G1 and pushes a cart carrying up to \SI{30}{kg} at \SI{0.5}{m/s}.
Future work will extend ForPC to other wheeled objects and contact modes and incorporate richer exteroception for transport in cluttered environments.

\bibliographystyle{elsarticle-num}
\bibliography{references}

\end{document}